\documentclass[letterpaper]{article} 
\usepackage{aaai2027}  
\usepackage[hyphens]{url}  
\usepackage{graphicx} 
\usepackage{natbib}  
\usepackage{caption} 
\usepackage{algorithm}
\usepackage{algorithmic}

\usepackage{newfloat}
\usepackage{listings}
\DeclareCaptionStyle{ruled}{labelfont=normalfont,labelsep=colon,strut=off} 
\floatstyle{ruled}
\newfloat{listing}{tb}{lst}{}
\floatname{listing}{Listing}

\usepackage{booktabs}

\usepackage{mathtools}
\usepackage{amssymb}
\usepackage[table]{xcolor}
\usepackage{multirow}
\usepackage{pifont}
\usepackage{subcaption}

\nocopyright

\title{TEA-AgriVLN: Traversability Estimation Alarm for Agricultural Vision-and-Language Navigation}
\author {
    Xiaobei Zhao\textsuperscript{\rm 1,\rm 2},
    Xingqi Lyu\textsuperscript{\rm 1},
    Xin Chen\textsuperscript{\rm 1,\rm 2}\corresponding,
    Xiang Li\textsuperscript{\rm 1,\rm 2}\corresponding
}
\affiliations {
    \textsuperscript{\rm 1}China Agricultural University\\
    \textsuperscript{\rm 2}China Agricultural University-Sichuan Advanced Agricultural \& Industrial Institute\\
    \{xiaobeizhao2002,lxq99725\}@163.com, \{chxin,cqlixiang\}@cau.edu.cn
}

\begin{document}

\maketitle

\begin{abstract}
Vision-and-Language Navigation in Continuous Environments (VLN-CE) requires an agent to follow a natural language instruction, predicting a sequence of low-level actions to navigate a robot from a starting point to a target location. The A2A benchmark and the AgriVLN method pioneeringly extended VLN-CE from indoor scenes to agricultural scenes, while we observed a challenging distinction: In indoor scenes, whether a zone is traversable tends to be clear to classify, such as wood floors are traversable but concrete walls are not. In agricultural scenes, however, this issue tends to be ambiguous, such as an unripe cornfield might be traversable for a robotic dog but might be non-traversable for a human. To address this issue, we propose the TEA module, which estimates the traversability of the camera image, then alarm the decision-maker for rethinking when the predicted action does not align with the traversability map. We integrate it into the AgriVLN backbone to build our TEA-AgriVLN method. When evaluated on A2A, it improves Success Rate (SR) from 0.47 to 0.54 and Navigation Error (NE) from 2.91 m to 2.70 m, showing the state-of-the-art performance in the agricultural VLN-CE domain. We further implement the ablation studies and the case study, discussing the effectiveness and limitations of TEA on different ground categories and scene classes.
\end{abstract}

\begin{links}
    \link{Code}{https://github.com/AlexTraveling/TEA-AgriVLN}
\end{links}

\begin{figure}[!t]
\centering
  \begin{subfigure}[t]{1.0\linewidth}
    \includegraphics[width=\linewidth]{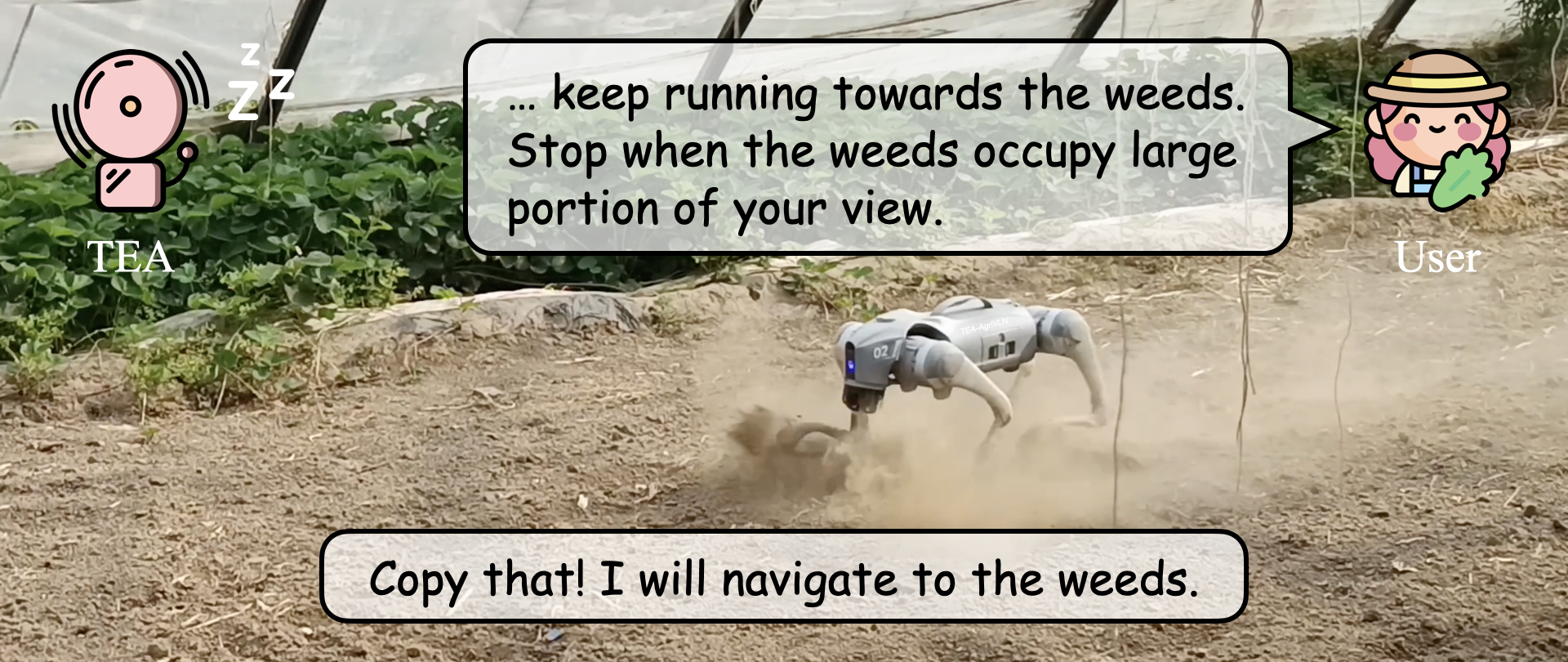}
  \end{subfigure}
  \hfill
  \vspace{-0.2cm}
  \begin{subfigure}[t]{1.0\linewidth}
    \includegraphics[width=\linewidth]{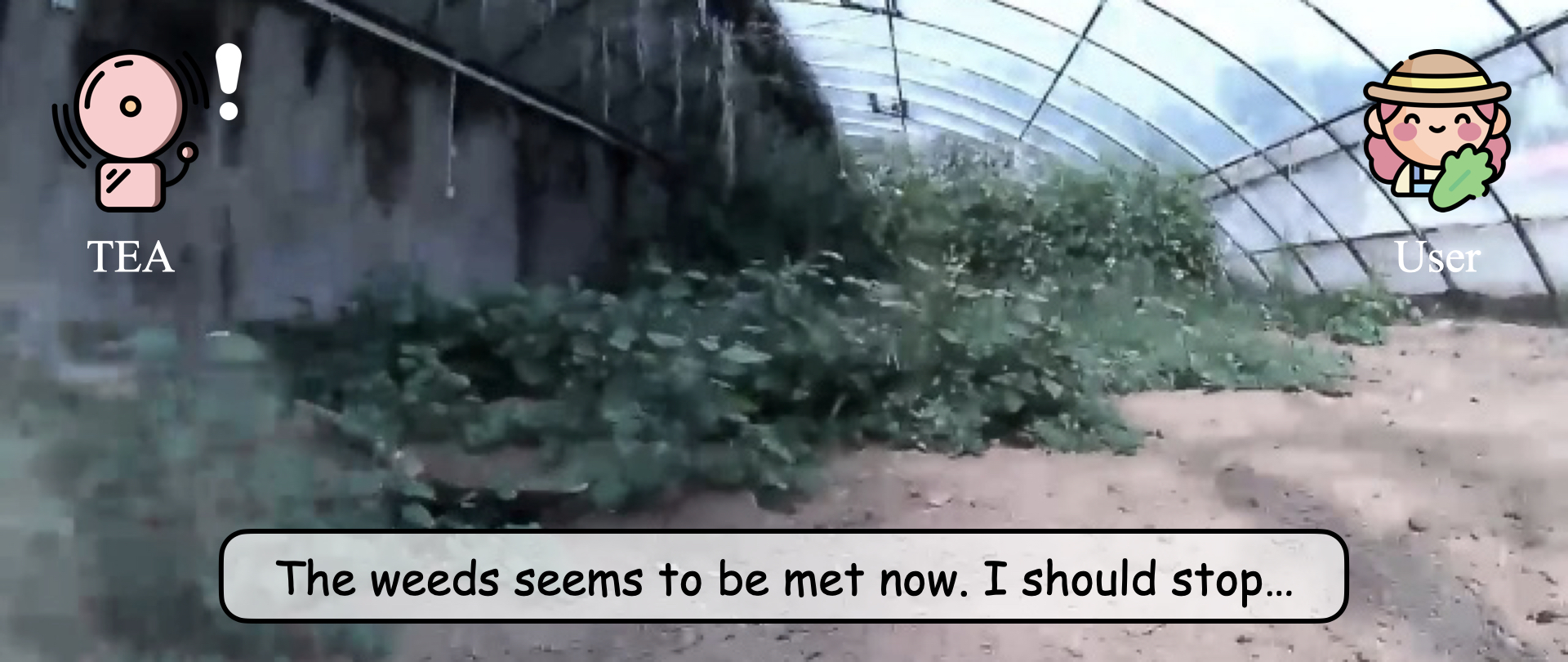}
  \end{subfigure}
  \hfill
  \vspace{-0.2cm}
  \begin{subfigure}[t]{1.0\linewidth}
    \includegraphics[width=\linewidth]{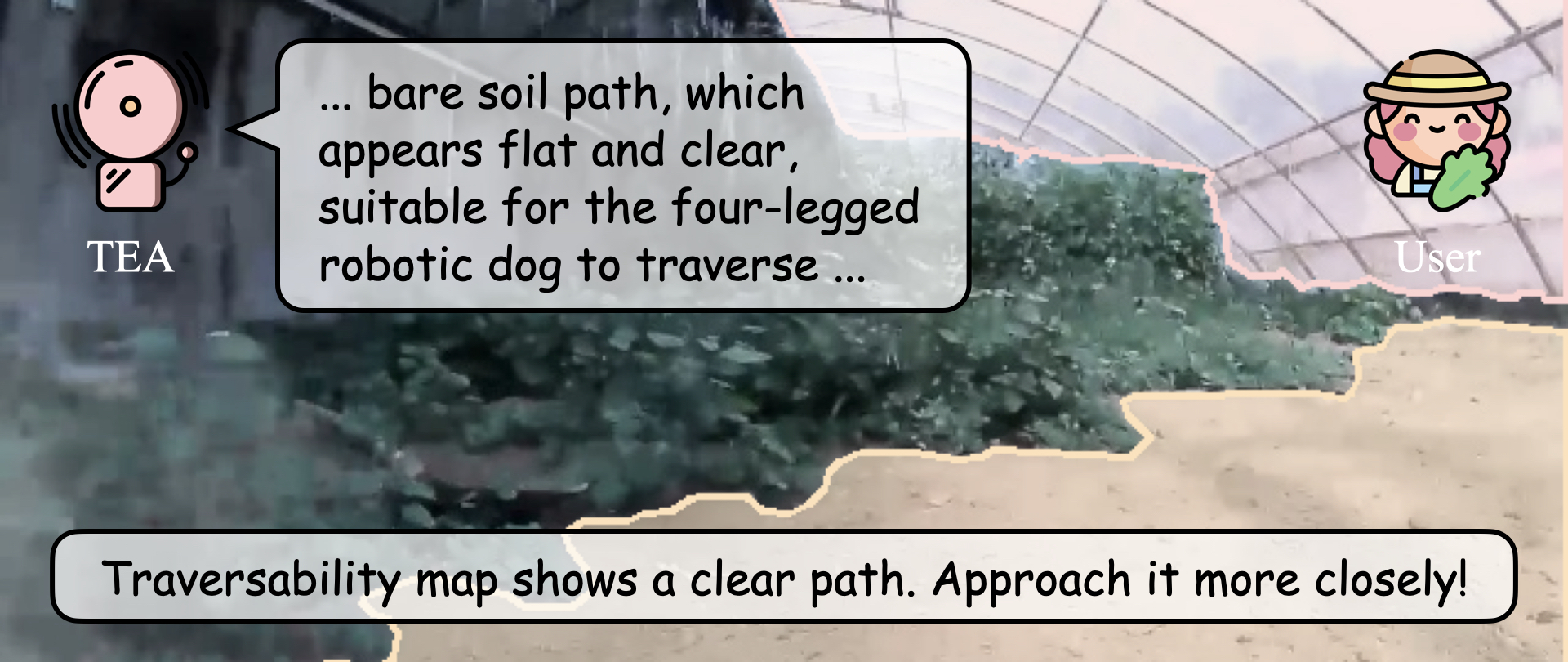}
  \end{subfigure}
\caption{TEA-AgriVLN running on a simple example.}
\label{fig:teaser}
\vspace{-0.4cm}
\end{figure}

\section{Introduction}

\par Agricultural robotic agents have been becoming useful helpers in a wide range of agricultural tasks, such as 
laser weeding \cite{COMPAG:LaserWeeding}, growth monitoring \cite{RAL:GrowthMonitoring} and cross-pollination \cite{Cell:Pollination}. 
However, most of them still heavily rely on manual operations or fixed railways for movement, resulting in limited mobility and poor adaptability on diversified scenarios.

\par In contrast, Vision-and-Language Navigation (VLN) \cite{CVPR:R2R} enables agents to follow the natural language instructions to navigate to the target positions, having demonstrated strong performance across various domains \cite{TMLR}, such as R2R \cite{CVPR:R2R} for indoor rooms, TOUCHDOWN \cite{CVPR:TOUCHDOWN} for urban streets, and AerialVLN \cite{ICCV:AerialVLN} for aerial spaces. 
To better align with the realistic conditions, 
Vision-and-Language Navigation in Continuous Environments (VLN-CE) \cite{ECCV:VLN-CE} further extends VLN from discrete topologies to continuous sequences. 
Motivated by prior Vision-Language Model (VLM)-based methods \cite{AAAI:NavGPT,ECCV:NavGPT-2,CVPR:Long}, Zhao et al. \cite{arXiv:AgriVLN} proposed the A2A benchmark and the AgriVLN method, pioneeringly extending VLN-CE from urban domains to agricultural domains.
When evaluated on A2A, however, AgriVLN only achieves Success Rate (SR) of 0.47 and Navigation Error (NE) of 2.91 m, leaving a large gap to the performance of human.


\par We observed a challenging distinction between agricultural scenes and urban scenes: In urban scenes, whether a zone is traversable tends to be clear to classify, such as wood floors are traversable but concrete walls are not. In agricultural scenes, however, this issue tends to be ambiguous, such as an unripe cornfield might be traversable for a robotic dog but might be non-traversable for a human.
This challenging distinction raised us a question: 
\textit{\textbf{May the ambiguous traversability be one of the reasons to the gap between AgriVLN and human?}}

\par To answer this question, we attempt to introduce a traversability estimator into the AgriVLN backbone. 
We approximately summarize the existing traversability estimation methods into two categories: the rule-based and the learning-based. 
Traditionally, the rule-based estimators \cite{CASE,RAL:GA-Nav,ICRA:Gaussian} use RGB camera to capture semantic features or depth camera to capture geometric features, then formulate a classification policy in a manual manner, which is good with efficiency and interpretability but is limited on poor generalization. 
Recently, the learning-based estimators \cite{RSS:FTE,ICRA:STEPP,ICRA:GSAT} use visual sensors to record environmental views and pose sensors to record running statues, then train a classification network in a self-supervised manner, which is good with accuracy and generalization but is dependent on elaborate training. 
Considering the complex semantic, the diversified geometric, and the limited affordance in agricultural scenes, we suggest that both rule-based and learning-based traversability estimators cannot satisfy our needs.


\par In this paper, we propose the module of 
\textbf{T}raversability 
\textbf{E}stimation 
\textbf{A}larm 
(\textbf{TEA}), which estimates the traversability of the camera image, then alarm the decision-maker for rethinking when the predicted action does not align with the traversability map. We integrate it into the AgriVLN backbone to build the method of 
\textbf{T}raversability 
\textbf{E}stimation 
\textbf{A}larm for 
\textbf{Agri}cultural 
\textbf{V}ision-and-\textbf{L}anguage \textbf{N}avigation 
(\textbf{TEA-AgriVLN}). 
When evaluated on A2A, our TEA-AgriVLN method improves SR from 0.47 to 0.54 and NE from 2.91 m to 2.70 m, showing the state-of-the-art performance in the agricultural VLN-CE domain. We further implement the ablation studies and the case study, discussing the effectiveness and limitations of TEA on different ground categories and scene classes.

\par Here we share a simple example to demonstrate how our TEA-AgriVLN method works, as illustrated in Figure \ref{fig:teaser}. 
After moving for a while, the decision-maker thinks that 
\textit{"the weeds seems to be met now"}, so improperly chooses to \texttt{[STOP]}. Immediately, the TEA module is activated. It thinks that \textit{"the flat and clear soil path is suitable for the four-legged robotic dog"}, so judges the alarming zone as traversable. Next, it alarms the decision-maker with the traversability map, successfully assisting it to rethink that 
\textit{"the traversability map indicates a clear path forward"} then properly change to choose to move \texttt{[FORWARD]}.

In summary, our main contributions are as follows:
\begin{itemize}
\item \textbf{TEA}, a traversability estimation-driven module, which can alarm the decision-maker for rethinking when the predicted action does not align with the traversability map.
\item \textbf{TEA-AgriVLN}, an agricultural VLN-CE method integrating the TEA module into the AgriVLN backbone, which can navigate agricultural robots to target positions following natural language instructions.
\item We implement the comparison experiment showing the state-of-the-art performance of TEA-AgriVLN in the agricultural VLN-CE domain, and we implement the ablation studies and the case study discussing the effectiveness and limitations of TEA in different conditions.
\end{itemize}




\section{Related Works}

\subsection{Benchmark for Agricultural VLN-CE}
\par Agriculture-to-Agriculture (A2A) \cite{arXiv:AgriVLN} is currently the only one VLN benchmark specially designed for agricultural robots, consisting of 1,560 episodes across 6 types of scene: farm, greenhouse, forest, mountain, garden and village, in which all the instructions belong to the step-by-step format and the action space belongs to the continuous environment.

\subsection{Method for Agricultural VLN-CE}
Vision-and-Language Navigation for Agricultural Robots (AgriVLN) \cite{arXiv:AgriVLN} is the first agricultural VLN-CE method, which uses NavGPT \cite{AAAI:NavGPT} as the backbone and integrate the Large Language Model (LLM)-based Subtask List (STL) module, enabling an agricultural robot navigating to the target positions following the natural language instructions.

\subsection{Traversability Estimation in Navigation}

\par In the macroscopic field of navigation, several studies have introduced traversability estimation to assist off-road navigation, 
such as FtF \cite{ECMR:FtF} in vegetations, GND \cite{ICRA:GND} in campuses, and V-STRONG \cite{ICRA:V-STRONG} in mountains. 
These studies show that traversability estimation can effectively assist decision-makers to better understand the environments.
In the specialized field of VLN, none of the existing studies, as far as we know, has done that before. 
The possible reason is that most scenes in classical VLN benchmarks belong to on-road scenes, in which whether a zone is traversable tends to be clear to classify. In the more specialized field of Agricultural VLN-CE, however, most scenes belong to off-road scenes. Hence, we suggest traversability estimation as a potential solution to the traverse ambiguity.


\begin{figure*}[t]
\centering
\includegraphics[width=1.0\linewidth]{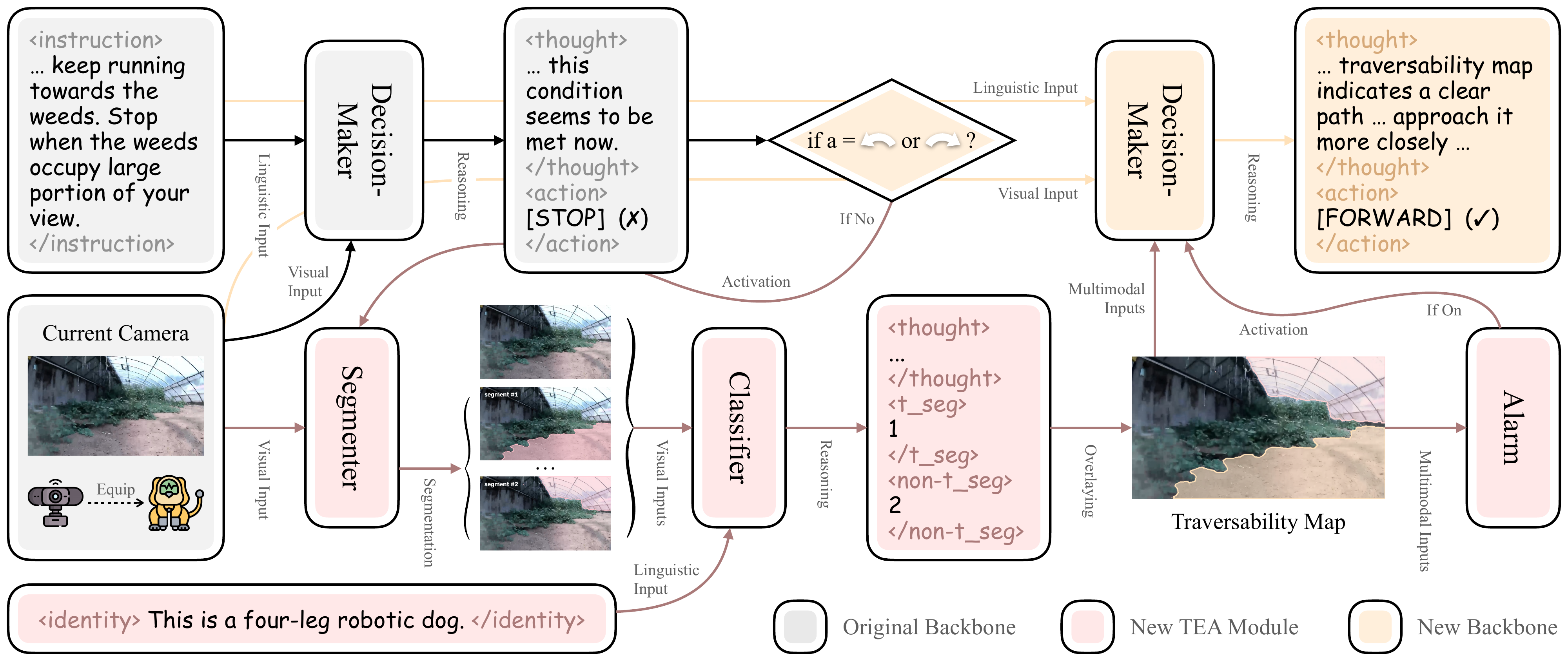}
\caption{TEA-AgriVLN illustration: 
The TEA module estimates the traversability of the camera image, then alarm the decision-maker for rethinking when the predicted action does not align with the traversability map. Please note that the two decision-makers are identical.
}
\label{fig:method}
\end{figure*}



\section{Methodology}
In this section, we present our TEA-AgriVLN method, 
as illustrated in Figure \ref{fig:method}. 
First, we introduce the task definition in Section \ref{sec:task_definition}. 
Second, we present our TEA module in Section \ref{sec:tea}. 
Third, we integrate the TEA module into the AgriVLN backbone to build the TEA-AgriVLN method in Section \ref{sec:backbone}.

\subsection{Task Definition}
\label{sec:task_definition}
The task of Agricultural Vision-and-Language Navigation \cite{arXiv:AgriVLN} is defined as follows: In each episode, the model is given an instruction in natural language, denoted as $W = \langle w_1, w_2, \dots, w_L \rangle$, where $L$ is the number of words. At each time step $t$, the model is given the front-facing RGB image $I_t$. The purpose is understanding both $W$ and $I_t$, to select the best low-level action $\hat{a_t}$ from action space $\{ \texttt{[FORWARD]}$, $\texttt{[LEFT ROTATE]}$, $\texttt{[RIGHT ROTATE]}$, $\texttt{[STOP]} \}$, thereby leading the robot navigate from the starting point to the target position. 

\subsection{TEA Module}
\label{sec:tea}
\par We design the TEA module running in a three-stage paradigm: segmentation, classification, and alarm.

\subsubsection{Stage One}
\par The inputs for the TEA module are an RGB image $I \in R^{3 \times h \times w}$ and a predicted action $\hat{a}$. 
We use a pre-trained instance segmenter 
$\mathcal{S}\,(\,\cdot\,)$ to divide $I$ into a group of small segments, defined as:
\begin{equation}
\left\{ M_1, M_2, M_3, \dots, M_m \right\} = \mathcal{S}(I)
\end{equation}
where $M_i \in \left\{ \texttt{True}, \texttt{False} \right\}^{h \times w}$ is the $i$-th segment. We calculate the area of $M_i$ by counting the number of pixel with the \texttt{True} value, then filter all the noisy segments with areas less than $\tau_a \times h \times w$, defined as:
\begin{equation}
\left\{ M_1, M_2, \dots, M_{m'} \right\} 
\xleftarrow[\tau_a]{\text{filter}}
\left\{ M_1, M_2, M_3, \dots, M_m \right\}
\end{equation}
where $\tau_a$ is the threshold of area ratio, and $m'$ is the length of the segment group after filtering.
We separately overlay every mask $M_i$ with its number $i$ on the original image $I$, defined as:
\begin{equation}
\left\{ O_1, O_2, \dots, O_{m'} \right\} 
\xleftarrow[c_u]{\text{overlay}}
I, \left\{ M_1, M_2, \dots, M_{m'} \right\} 
\end{equation}
where $c_u$ is the color mode for overlaying unclassified mask. 
$O_i \in R^{3 \times h \times w}$ is the $i$-th overlaid image, in which we overlay the mask on the corresponding position and overlay the number on the top-left corner.


\subsubsection{Stage Two}
We prompt a pre-trained 
VLM 
in a zero-shot manner as the classifier $\mathcal{C}\,(\,\cdot\,)$, which takes the robot identity $R$, the camera image $I$, and the group of overlaid images $\left\{ O_1, O_2, \dots, O_{m'} \right\}$ as the inputs, 
to classify whether a mask overlaid area is traversable, defined as:
\begin{equation}
T 
= 
\mathcal{C}\big(P_{sys}^{\mathcal{C}}, P_{usr}^{\mathcal{C}}(R, I, \left\{ O_1, O_2, \dots, O_{m'} \right\})\big)
\end{equation}
where $P_{sys}^{\mathcal{C}}$ and $P_{usr}^{\mathcal{C}}\,(\,\cdot\,)$ are the system and user prompt templates for $\mathcal{C}\,(\,\cdot\,)$, respectively. 
$T$ is the textual response message. 
We design $P_{sys}^{\mathcal{C}}$ in the following format:

\begin{quote}
\textit{You are an expert in agricultural robot navigation. Your task is to understand both the robot identity and the scene images, to classify each candidate segment as either traversable or non-traversable for this robot.\\ \\
Input:\\
<robot\_identity> ... </robot\_identity>\\
<images> ... </images>\\ \\
Output format:\\ 
<thought\_summary> ... </thought\_summary>\\
<traversable\_seg> ... </traversable\_seg>\\
<non-traversable\_seg> ... </non-traversable\_seg>
}
\end{quote}
Then we use the simple yet effective regex match to extract the three tag pairs from $T$, defined as:
\begin{equation}
\rho_{\mathcal{C}}, G_+, G_-
\xleftarrow[]{\text{extract}}
T
\end{equation}
where $\rho_{\mathcal{C}}$ is the thought summary of $\mathcal{C}\,(\,\cdot\,)$, which provides an explicit interpretation to its reasoning process. 
$G_+$ and $G_-$ are the sets of the numbers of the traversable and non-traversable regions, respectively. 
$G_+ \cup G_- = \left\{ 1, 2, \dots, m' \right\}$ and $G_+ \cap G_- = \varnothing$.
We divide the set of candidate regions $\left\{ M_1, M_2, \dots, M_{m'} \right\}$ into the set of traversable regions $\left\{ M_i \mid i \in G_+ \right\}$ and the set of non-traversable regions $\left\{ M_j \mid j \in G_- \right\}$, which serves as the output of traversability estimation. We overlay them together on the original camera image $I$, defined as:
\begin{equation}
\label{equ:traversability_map}
O
\xleftarrow[c_+, c_-]{\text{overlay}}
I, \left\{ M_i \mid i \in G_+ \right\}, \left\{ M_j \mid j \in G_- \right\}
\end{equation}
where $c_+$ and $c_-$ are the different color modes for overlaying traversable and non-traversable masks. 
$O \in R^{3 \times h \times w}$ is the traversability map in the RGB representation.

\subsubsection{Stage Three}
If a robot was moving forward in a 3D environment, there would be a zone in front of it that it is about to pass through. In this paper, we define this zone as the alarming zone. 
When we map it onto a 2D camera image, it transforms into a trapezoidal shape, so we denote the alarming zone on 2D camera images as $Z = (t_z, b_z, h_z)$, where $t_z$, $b_z$ and $h_z$ are the top base ratio, bottom base ratio, and height ratio, respectively.

We respectively calculate the intersection areas of the traversable regions $\left\{ M_i \mid i \in G_+ \right\}$ and the non-traversable regions $\left\{ M_j \mid j \in G_- \right\}$ inside the alarming zone $Z$, defined as:
\begin{equation}
S'_{+}
=
\left|
\left( \bigcup_{i\in G_+}M_i \right)
\cap
Z
\right|
\end{equation}
where $\left|\,\cdot\,\right|$ is the quantity of points. Then we calculate the proportion of the intersection area $S'_{+}$ in the alarming zone $Z$, defined as:
\begin{equation}
S_{+} = \frac{S'_{+}}{\left| Z \right|}.
\end{equation}
where $S_{+}$ is the intersection proportion of the traversable regions, and we calculate the intersection proportion of the non-traversable regions $S_{-}$ in the same paradigm.

\par Next, according to the predicted action $\hat{a}$, we design the rule-based alarm $\mathcal{A}\,(\,\cdot\,)$ running in a bi-directional activation manner:

\begin{itemize}
\item If $\hat{a} = \texttt{[FORWARD]}$, the goal is to judge whether the alarming zone includes non-traversable regions, then alarm the decision-maker that moving forward might be dangerous when happened, defined as:
\begin{equation}
\label{equ:forward}
\mathcal{A}(S_+, S_-)=
\begin{cases}
\texttt{on}, & \text{if } S_- > \tau_-,\\
\texttt{off}, & \text{otherwise.}
\end{cases}
\end{equation}
\item If $\hat{a} = \texttt{[STOP]}$, the goal is to judge whether the alarming zone has both enough traversable regions and few enough non-traversable regions, then alarm the decision-maker that stop moving might be conservative when happened, defined as:
\begin{equation}
\label{equ:stop}
\mathcal{A}(S_+, S_-)=
\begin{cases}
\texttt{on}, & \text{if } S_+ > \tau_+ \land S_- < \tau_-,\\
\texttt{off}, & \text{otherwise.}
\end{cases}
\end{equation}
\end{itemize}
where $\tau_+$ and $\tau_-$ are the thresholds of traversable and non-traversable intersection proportions, respectively. 
$A \in \left\{ \texttt{on}, \texttt{off} \right\}$
serves as the output of the TEA module.



\subsection{Backbone}
\label{sec:backbone}
\par To build the TEA-AgriVLN method, we follow the architecture of AgriVLN \cite{arXiv:AgriVLN} as our backbone, and we integrate the TEA module as a post-processing layer after the decision-making. 

\par We prompt a pre-trained VLM in a one-shot manner as the decision-maker $\mathcal{D}\,(\,\cdot\,)$.
At each time step $t$, $\mathcal{D}\,(\,\cdot\,)$ takes the instruction $W$ and the current camera image $I_t$ as the inputs, to predict the most appropriate low-level action $\hat{a_t}$, defined as:
\begin{equation}
T
= 
\mathcal{D}\big(P_{sys}^{\mathcal{D}}, P_{usr}^{\mathcal{D}}(\text{STL}(W), I_t)\big)
\end{equation}
\begin{equation}
\hat{a_{t}}, \rho_t 
\xleftarrow[]{\text{extract}}
T
\end{equation}
where $\text{STL}\,(\,\cdot\,)$ is the Subtask List module followed from the AgriVLN backbone. $P_{sys}^{\mathcal{D}}$ and $P_{usr}^{\mathcal{D}}\,(\,\cdot\,)$ are the system and user prompt templates for $\mathcal{D}\,(\,\cdot\,)$, respectively. $\rho_{t}$ is the thought summary of $\mathcal{D}\,(\,\cdot\,)$, which provides an explicit interpretation to its reasoning process. 

\par When $\hat{a_{t}} \in \left\{ \texttt{[LEFT ROTATE]}, \texttt{[RIGHT ROTATE]} \right\}$, TEA keeps silent. 
When $\hat{a_{t}} \in \left\{ \texttt{[FORWARD]}, \texttt{[STOP]} \right\}$, TEA is activated to estimate the traversability of $I_t$ then raise the alarm according to $\hat{a_{t}}$, defined as:
\begin{equation}
A_t, O_t
= 
\text{TEA} (I_t, \hat{a_{t}})
\end{equation}
where $O_t \in R^{3 \times h \times w}$ is the traversability map from Equation \ref{equ:traversability_map}, and $A_t \in \left\{ \texttt{on}, \texttt{off} \right\}$ is the alarm result from Equation \ref{equ:forward} or \ref{equ:stop}. 
If and only if $A_t$ = \texttt{on}, we re-prompt the identical decision-maker $\mathcal{D}\,(\,\cdot\,)$. 
In this turn, we additionally input the traversability map $O_t$, and insert a sentence to describe it in the system prompt $P_{sys}^{\mathcal{D}}$, defined as:
\begin{equation}
\underline{T}
= 
\mathcal{D}\big(\underline{P_{sys}^{\mathcal{D}}}, P_{usr}^{\mathcal{D}}(\text{STL}(W), I_t, \underline{O_t})\big)
\end{equation}
\begin{equation}
\underline{\hat{a_{t}}}, \underline{\rho_t} 
\xleftarrow[]{\text{extract}}
\underline{T}
\end{equation}
where \underline{underline} marks the differences between the two decision-making turns. We use the new predicted action $\hat{a_{t}}$ to replace the old one.

\par As the time step $t$ goes by, the predicted action sequence $\langle \hat{a_{0}}, \hat{a_{1}}, \hat{a_{2}}, \dots, \hat{a_{t}} \rangle$ navigates the robot from the starting point to the target position. 
In an episode, $\mathcal{D}\,(\,\cdot\,)$ ends when one of the following conditions happens: 
\begin{quote}
1) $\hat{a_{t'}}$ = $\texttt{[STOP]}$; \\
2) The predicted action sequence $\langle \hat{a_{t'-\delta}}, \hat{a_{t'-\delta+1}},$ $\dots, \hat{a_{t'}} \rangle$ is deviated to the ground-truth action sequence $\langle a_{t'-\delta}, a_{t'-\delta+1}, \dots, a_{t'} \rangle$; \\
3) $t'$ reaches the max limitation of time step.
\end{quote}
where $\delta$ is the deviation time threshold. $t'$ is the ending time step.

\begin{table*}[t]

\begin{center}
\renewcommand{\arraystretch}{1.2}
\begin{tabular}{rllcc|cc|cccc}
\toprule
\multirow{2}{*}{\textbf{\#}} & \multirow{2}{*}{\textbf{Method}} & \multirow{2}{*}{\textbf{Publication}} & 
\multicolumn{2}{c}{\textbf{A2A (low)}} & 
\multicolumn{2}{c}{\textbf{A2A (high)}} & 
\multicolumn{2}{c}{\textbf{A2A}} \\
\cmidrule(lr){4-5} \cmidrule(lr){6-7} \cmidrule(lr){8-9}
& & & \textbf{SR} $\uparrow$ & \textbf{NE} $\downarrow$     &
\textbf{SR} $\uparrow$ & \textbf{NE} $\downarrow$ & 
\textbf{SR} $\uparrow$ & \textbf{NE} $\downarrow$ \\
\midrule
1 & Random & - & 0.13 & 7.30 & 0.04 & 6.74 & 0.09 & 7.03 \\
2 & Fixed  & - & 0.00 & 0.00 & 0.06 & 6.32 & 0.03 & 3.06 \\
\midrule 
3 & SIA-VLN \cite{EMNLP:SIA-VLN}   & EMNLP 2020 & 0.52             & 1.46              & \underline{0.08} & \underline{5.12} & \underline{0.31} & \underline{3.24} \\
4 & NavGPT \cite{AAAI:NavGPT}      & AAAI 2024  & 0.51             & \textbf{0.60}     & 0.14             & 5.01             & 0.33             & 2.76             \\
5 & DILLM-VLN \cite{RAL:DILLM-VLN} & RA-L 2025  & \underline{0.41} & 1.36              & 0.32             & 3.90             & 0.36             & \textbf{2.60}    \\
6 & AgriVLN \cite{arXiv:AgriVLN}   & arXiv 2025 & 0.58             & \underline{2.32}  & 0.35             & 3.54             & 0.47             & 2.91             \\
\rowcolor{gray!15}
7 & TEA-AgriVLN (Ours)             & -          & \textbf{0.65}    & 2.09              & \textbf{0.43}    & \textbf{3.36}    & \textbf{0.54}    & 2.70             \\
\midrule
8 & Human  & - & 0.93 & 0.32 & 0.80 & 0.82 & 0.87 & 0.57   \\
\bottomrule
\end{tabular}
\end{center}

\caption{Comparison results between our TEA-AgriVLN method and the state-of-the-art methods on the low-complexity portion (subtask $=$ 2), the high-complexity portion (subtask $\geq$ 3), and the whole of the A2A benchmark. 
\textbf{Bold} and \underline{underline} mark the best and worst scores, respectively.
}
\label{tab:comparison_experiment}

\end{table*}

\section{Experiments}

\subsection{Experimental Settings}
\label{sec:experimental_settings}
We implement all the experiments on the A2A \cite{arXiv:AgriVLN} benchmark. 
Following the settings in A2A, we adopt GPT-4.1 as the LLM and GPT-4.1-mini as the 
VLM.
We access all the LLMs and VLMs through APIs. 
Besides, we locally deploy pre-trained SAM2.1-large as the instance segmenter. 
Regarding hyper-parameters, we set the image size $h$ = 360 and $w$ = 640, 
the filtering area ratio threshold $\tau_a$ = 0.02, 
the overlaying color modes in BGR $c_u$ = (210, 210, 255), $c_+$ = (185, 225, 255) and $c_-$ = (210, 210, 255), 
the robot identity $R$ = \textit{"This is a four-leg robotic dog."}, 
the alarming zone size $t_z$ = 0.32, $b_z$ = 0.4 and $h_z$ = 0.1, 
the inference temperature = 1e-4, 
the intersection proportion thresholds $\tau_+$ = 0.5 and $\tau_-$ = 0.1, 
and the deviation time threshold $\delta$ = 4 s.
All the experiments run on a single NVIDIA L20 GPU with 48 G video memory.



\subsection{Evaluation Metrics}
\label{sec:evaluation_metrics}
We follow the two standard VLN evaluation metrics \cite{CVPR:R2R}: Success Rate (SR) and Navigation Error (NE). NE measures the path length between the stopping position and the target position. SR measures the rate successfully reaching the target position within a 2-meter NE. 
We consider SR as the primary evaluation metric.


\subsection{Comparison Experiment}
\label{sec:comparison_experiment}
We compare our TEA-AgriVLN method with four state-of-the-art methods on the A2A benchmark. 
Besides, we follow the scores of the methods of Random, Fixed and Human as the lower and upper bounds, respectively.
\par The comparison experiment results are shown in Table \ref{tab:comparison_experiment}. On the whole of A2A, our TEA-AgriVLN method achieves the SR of 0.54 and the NE of 2.70 m, successfully increasing SR by 7 percentage points and decreasing NE by 0.21 m compared to the AgriVLN baseline. 
Comprehensively considering the best SR and the second best NE, 
we think that our TEA-AgriVLN method achieves the state-of-the-art performance in the agricultural VLN domain.
\par From this comparison, we suggest that the TEA module can effectively estimate the traversability of the camera image, and alarm the decision-maker for rethinking when the predicted action does not align with the traversability map.



\subsection{Ablation Studies}
\label{sec:ablation_study}

\par To further study the effect of our TEA module, we implement the ablation studies on four aspects: its three stages, task complexities, ground categories, and scene classes.



\subsubsection{Three Stages in TEA} 
\par We ablate the three stages in our TEA module. 
For ablating the segmentation stage, we replace the segmentation masks with the alarming zone. For ablating the classification stage, we remove the robot identity. The alarming stage is necessary, so we always keep it.
\par As shown in Table \ref{tab:ablation_three_stages}, 
when the segmentation stage is ablated, SR drops by 2 percentages and NE drops by 0.05 m, from which we suggest that the traversability information out of the alarming zone is also valuable for the decision-maker. When the classification stage is ablated, SR drops by 3 percentages and NE drops by 0.08 m, from which we suggest that the robot identity is valuable for the classifier to build the correct self-cognition on trafficability. 
In summary, we suggest that both stages are valuable, and they collaborate in a compatible manner.



\begin{table}[!h]
\centering
\renewcommand{\arraystretch}{1.2}
\begin{tabular}{rccccc}
\toprule
\multirow{2}{*}{\#} &
\multicolumn{3}{c}{\textbf{TEA}} &
\multirow{2}{*}{\textbf{SR} $\uparrow$} &
\multirow{2}{*}{\textbf{NE} $\downarrow$} \\
\cmidrule(lr){2-4}
& \textbf{Segm.} & \textbf{Clas.} & \textbf{Alrm.} & & \\
\midrule

\rowcolor{gray!15}
9  & \ding{51} & \ding{51} & \ding{51} & \textbf{0.54}    & \textbf{2.70} \\
10 & \ding{55} & \ding{51} & \ding{51} & 0.52             & 2.75 \\
11 & \ding{51} & \ding{55} & \ding{51} & 0.51             & 2.78 \\
12 & \ding{55} & \ding{55} & \ding{55} & \underline{0.47} & \underline{2.91} \\

\bottomrule
\end{tabular}

\caption{Ablation results on the three stages in TEA. 
\textbf{Bold} and \underline{underline} mark the best and worst scores, respectively.
}
\label{tab:ablation_three_stages}

\end{table}

\subsubsection{Task Complexity} 
We follow the subtask quantity distribution in the A2A benchmark.
As shown in Table \ref{tab:ablation_task_complexity}, when the TEA module is removed, we observe that SRs and NEs consistently drop across different task complexities. 
\par From this ablation study, we suggest that our TEA module works consistently across different task complexities, and
we can eliminate this factor in the following discussions. 



\begin{table}[h]

\centering
\renewcommand{\arraystretch}{1.2}
\begin{tabular}{rcccc}
\toprule
\# & \textbf{Subtask} & \textbf{Method} & \textbf{SR} $\uparrow$ & \textbf{NE} $\downarrow$ \\ 
\midrule

13 & \multirow{2}{*}{2}        & Ours     & 0.65 & 2.09 \\
14 &                           & w/o TEA  & 0.57 & 2.32 \\
\rowcolor{gray!15}
\multicolumn{3}{c}{$\Delta$} & - 0.08 & + 0.23 \\
\midrule
15 & \multirow{2}{*}{$\geq$ 3} & Ours     & 0.43 & 3.36 \\
16 &                           & w/o TEA  & 0.35 & 3.54 \\
\rowcolor{gray!15}
\multicolumn{3}{c}{$\Delta$} & - 0.08 & + 0.18 \\

\bottomrule
\end{tabular}

\caption{Ablation results on task complexities. 
“$\Delta$” represents the score change after the TEA module is removed.
}
\label{tab:ablation_task_complexity}

\end{table}

\subsubsection{Ground Category} 
\label{sec:ablation_ground_category}
According to the ground categories of the robot moving trajectories, we manually partition the full A2A benchmark into four parts: paved roads, dirts, meadows, and sparse plants. 
\par As shown in Table \ref{tab:ablation_ground_category}, when the TEA module is removed, we observe the significant drops in paved roads and meadows, the major drop in dirts, and the slight drop in sparse plants. 
We attribute this large gap to the different semantic complexities across the different ground categories: In paved roads, such as a cement road, the semantic complexities tend to be single and pure. In sparse plants, such as an unripe cornfield, however, the semantic complexities tend to be multiple and chaotic. This semantic complexity distinction brings more challenges for the instance segmentation and the traversability classification, which results in the relatively poor qualities of the traversability maps.
\par From this ablation study, we suggest that our TEA module works effectively across different ground categories, and works better on paved roads and meadows. 

\begin{table}[!h]

\centering
\renewcommand{\arraystretch}{1.2}
\begin{tabular}{rcccc}
\toprule
\# & \textbf{Ground} & \textbf{Method} & \textbf{SR} $\uparrow$ & \textbf{NE} $\downarrow$ \\ 
\midrule
17 & \multirow{2}{*}{Paved Road}   & Ours    & \textbf{0.56} & \textbf{3.13} \\
18 &                               & w/o TEA & 0.43          & 3.34 \\
\midrule
19 & \multirow{2}{*}{Dirt}         & Ours    & \textbf{0.60} & \textbf{2.27} \\
20 &                               & w/o TEA & 0.57          & 2.33 \\
\midrule
21 & \multirow{2}{*}{Meadow}       & Ours    & \textbf{0.47} & \textbf{2.33} \\
22 &                               & w/o TEA & 0.40          & 2.76 \\
\midrule
23 & \multirow{2}{*}{Sparse Plant} & Ours    & \textbf{0.55} & \textbf{2.65} \\
24 &                               & w/o TEA & \textbf{0.55} & 2.74 \\
\bottomrule
\end{tabular}

\caption{Ablation results on ground categories. 
In every ground category, \textbf{bold} marks the better score.
}
\label{tab:ablation_ground_category}

\end{table}




\subsubsection{Scene Class} 
We follow the six scene classes in the A2A benchmark. As shown in Table \ref{tab:ablation_scene_class}, 
when the TEA module is removed, we observe the significant drops in villages and greenhouses, the major drops in farms, forests and gardens, and the unexpected slight improvements in mountains. We are surprised that our TEA-AgriVLN method almost catch up with the performance of Human in villages (\#8, \#35), but even slightly go backward in mountains (\#31, \#32). We attribute this large gap to the different ground categories across the different scene classes: In villages and greenhouses, plenty of grounds belong to paved roads. In mountains, however, most grounds belong to dirts, meadows and sparse plants. Clearly, this ablation result matches with the ablation result on ground categories.
\par From this ablation study, we suggest that our TEA module works effectively across most scene classes, and works better in villages and greenhouses.

\begin{table}[h]

\centering
\renewcommand{\arraystretch}{1.2}
\begin{tabular}{rcccc}
\toprule
\# & \textbf{Scene} & \textbf{Method} & \textbf{SR} $\uparrow$ & \textbf{NE} $\downarrow$ \\ 
\midrule

25 & \multirow{2}{*}{Farm}       & Ours     & \textbf{0.71} & \textbf{1.74} \\
26 &                             & w/o TEA  & 0.66          & 1.79 \\
\midrule
27 & \multirow{2}{*}{Greenhouse} & Ours     & \textbf{0.49} & \textbf{2.35} \\
28 &                             & w/o TEA  & 0.35          & 2.56 \\
\midrule
29 & \multirow{2}{*}{Forest}     & Ours     & \textbf{0.38} & \textbf{3.31} \\
30 &                             & w/o TEA  & 0.33          & 3.68 \\
\midrule
31 & \multirow{2}{*}{Mountain}   & Ours     & \textbf{0.50} & 2.85 \\
32 &                             & w/o TEA  & \textbf{0.50} & \textbf{2.84} \\
\midrule
33 & \multirow{2}{*}{Garden}     & Ours     & \textbf{0.56} & \textbf{3.67} \\
34&                             & w/o TEA  & 0.49          & 3.77 \\
\midrule
35& \multirow{2}{*}{Village}    & Ours     & \textbf{0.80} & \textbf{1.99} \\
36&                             & w/o TEA  & 0.47          & 2.92 \\

\bottomrule
\end{tabular}

\caption{Ablation results on scene classes. 
In every scene class, \textbf{bold} marks the better score.
}
\label{tab:ablation_scene_class}

\end{table}

\begin{figure*}[t]
\centering
\includegraphics[width=1.0\linewidth]{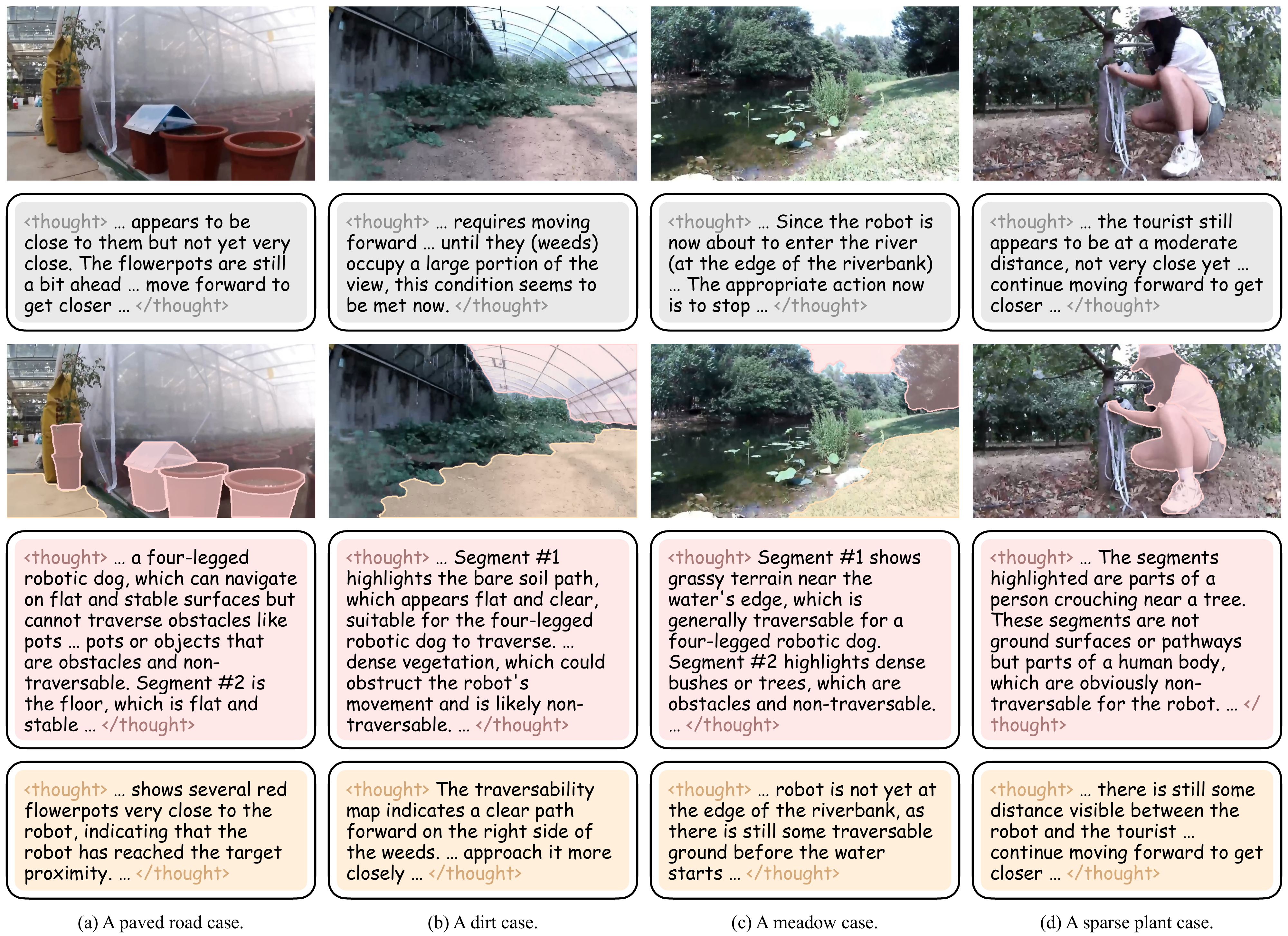}
\caption{Case study illustration: In every case, the images or textboxes from top to bottom are the camera image, the thinking thought on action prediction, the traversability map, the thinking thought on traversability classification, the rethinking thought on action prediction, respectively.}
\label{fig:case_study}
\end{figure*}


\subsection{Case Study}
\label{sec:case_study}
\par To make our case study be more comprehensive, we select three successful episodes and one failed episode from four different ground categories, as illustrated in Figure \ref{fig:case_study}. We mark all the pivotal reasoning thoughts with \textit{italics} in the text and with \underline{underline} in the figure.


\par Here we take the case running on a paved road as an example, to explain how our TEA module works. In this case, the goal is to reach the red sunflower pots. At the current time step, this goal is satisfied, so the ground-truth action should be \texttt{[STOP]}. However, the decision-maker improperly thinks that \textit{"the flowerpots are still a bit ahead"}, so improperly chooses to move \texttt{[FORWARD]}, which results in a collision (SR = 0, NE = 0.00 m). When the TEA module is integrated, considering that this robot is \textit{"a four-legged robotic dog"}, the TEA module thinks that it \textit{"can navigate on flat and stable surfaces but cannot traverse obstacles like pots"}, so classifies the floor as traversable and the pots as non-traversable. Next, the TEA module alarms the decision-maker with the traversability map, successfully assisting it to think that \textit{"the robot has reached the target proximity"} then properly change to choose \texttt{[STOP]} (SR = 1, NE = 0.62 m).

\par From the successful cases on a paved road, a dirt, and a meadow, we suggest that our TEA module shows the effective bi-directional alarming function:
When the decision-maker chose to move \texttt{[FORWARD]}, TEA could alarm it for rethinking if there were certain dangerous obstacles in the alarming zone; 
When the decision-maker chose to \texttt{[STOP]}, TEA could alarm it for rethinking if there were enough safe spaces in the alarming zone. 

\par Meanwhile, our TEA-AgriVLN method fails on the case in sparse plants. We observe that the TEA module correctly segments and classifies the person, but omits the tree. We attribute this omitting to the instance segmentation stage: The semantic similarity between the tree trunk and the fallen leaves is very high, which makes the pre-trained segmenter struggle for distinguishing them.
Clearly, this case study also matches with the ablation results on ground categories.

\par From the failed case in sparse plants, we suggest that 
the chaotic off-road grounds are still challenging for the TEA module to estimate traversability.

\section{Conclusion}
\par In this paper, we present the TEA module, 
which estimates the traversability of the camera image, then alarm the decision-maker for rethinking when the predicted action does not align with the traversability map.
We integrate it into the AgriVLN backbone to build our TEA-AgriVLN method, successfully improving SR from 0.47 to 0.54 and NE from 2.91 m to 2.70 m, 
from which 
we suggest an answer to the opening question: 
\textit{\textbf{Yes, the ambiguous traversability is one of the reasons to the gap between AgriVLN and human, and our TEA module can effectively narrow this gap.}}

\par During experiments, 
we also find a main limitation.
As discussed in Section \ref{sec:ablation_ground_category} and \ref{sec:case_study}, chaotic off-road grounds, such as sparse plants, are still challenging for our TEA module.

\par In the future, we plan to further explore 
if we could fine-tuned the instance segmenter, 
would our TEA module perform better on chaotic off-road grounds.

\bibliography{aaai2027}

@article{COMPAG:LaserWeeding,
  author  = "Peng Zhao and Junlin Chen and Jiahao Li and Jifeng Ning and Yongming Chang and Shuqin Yang",
  title   = "{Design and Testing of an autonomous laser weeding robot for strawberry fields based on DIN-LW-YOLO}",
  journal = "Computers and Electronics in Agriculture",
  year    = 2025,
}

@article{RAL:GrowthMonitoring,
  author  = "Luca Lobefaro and Matteo Sodano and Daniel Fusaro and Federico Magistri and Meher V. R. Malladi and Tiziano Guadagnino and Alberto Pretto and Cyrill Stachniss",
  title   = "{Spatio-Temporal Consistent Semantic Mapping for Robotics Fruit Growth Monitoring}",
  journal = "Robotics and Automation Letters",
  year    = 2025,
}

@article{Cell:Pollination,
  author  = "Yue Xie and Tinghao Zhang and Minghao Yang and Hongchang Lyu and Yupan Zou and Yangchang Sun and Jun Xiao and Wenzhao Lian and Jianhua Tao and Hua Han and Cao Xu",
  title   = "{Engineering crop flower morphology facilitates robotization of cross-pollination and speed breeding}",
  journal = "Cell",
  year    = 2025,
}

@InProceedings{CVPR:R2R,
  author    = "Peter Anderson and Qi Wu and Damien Teney and Jake Bruce and Mark Johnson and Niko Sunderhauf and Ian Reid and Stephen Gould and Anton van den Hengel",
  title     = "{Vision-and-Language Navigation: Interpreting visually-grounded navigation instructions in real environments}",
  booktitle = "IEEE/CVF Conference on Computer Vision and Pattern Recognition (CVPR)",
  year      = 2018,
}

@article{TMLR,
  author  = "Yue Zhang and Ziqiao Ma and Jialu Li and Yanyuan Qiao and Zun Wang and Joyce Chai and Qi Wu and Mohit Bansal and Parisa Kordjamshidi",
  title   = "{Vision-and-Language Navigation Today and Tomorrow: A Survey in the Era of Foundation Models}",
  journal = "Transactions on Machine Learning Research",
  year    = 2024,
}

@InProceedings{CVPR:TOUCHDOWN,
  author    = "Howard Chen and Alane Suhr and Dipendra Misra and Noah Snavely and Yoav Artzi",
  title     = "{TOUCHDOWN: Natural Language Navigation and Spatial Reasoning in Visual Street Environments}",
  booktitle = "IEEE/CVF Conference on Computer Vision and Pattern Recognition (CVPR)",
  year      = 2019,
}

@InProceedings{ICCV:AerialVLN,
  author    = "Shubo Liu and Hongsheng Zhang and Yuankai Qi and Peng Wang and Yanning Zhang and Qi Wu",
  title     = "{AerialVLN: Vision-and-Language Navigation for UAVs}",
  booktitle = "IEEE/CVF International Conference on Computer Vision (ICCV)",
  year      = 2023,
}

@InProceedings{ECCV:VLN-CE,
  author    = "Jacob Krantz and Erik Wijmans and Arjun Majumdar and Dhruv Batra and Stefan Lee",
  title     = "{Beyond the Nav-Graph: Vision-and-Language Navigation in Continuous Environments}",
  booktitle = "European Conference on Computer Vision (ECCV)",
  year      = 2020,
}

@InProceedings{AAAI:NavGPT,
  author    = "Gengze Zhou and Yicong Hong and Qi Wu",
  title     = "{NavGPT: Explicit Reasoning in Vision-and-Language Navigation with Large Language Models}",
  booktitle = "AAAI Conference on Artificial Intelligence (AAAI)",
  year      = 2024,
}

@InProceedings{ECCV:NavGPT-2,
  author    = "Gengze Zhou and Yicong Hong and Zun Wang and Xin Eric Wang and Qi Wu",
  title     = "{NavGPT-2: Unleashing Navigational Reasoning Capability for Large Vision-Language Models}",
  booktitle = "European Conference on Computer Vision (ECCV)",
  year      = 2024,
}

@InProceedings{CVPR:Long,
  author    = "Xinshuai Song and Weixing Chen and Yang Liu and Weikai Chen and Guanbin Li and Liang Lin",
  title     = "{Towards Long-Horizon Vision-Language Navigation: Platform, Benchmark and Method}",
  booktitle = "IEEE/CVF Conference on Computer Vision and Pattern Recognition (CVPR)",
  year      = 2025,
}

@InProceedings{arXiv:AgriVLN,
  author    = "Xiaobei Zhao and Xingqi Lyu and Xiang Li",
  title     = "{AgriVLN: Vision-and-Language Navigation for Agricultural Robots}",
  booktitle = "arXiv 2508.07406",
  year      = 2025,
}

@InProceedings{CASE,
  author    = "Hosseinpoor, Sadegh and Torresen, Jim and Mantelli, Mathias and Pitto, Diego and Kolberg, Mariana and Maffei, Renan and Prestes, Edson",
  title     = "{Traversability Analysis by Semantic Terrain Segmentation for Mobile Robots}",
  booktitle = "IEEE International Conference on Automation Science and Engineering (CASE)",
  year      = 2021,
}

@article{RAL:GA-Nav,
  author  = "Guan, Tianrui and Kothandaraman, Divya and Chandra, Rohan and Sathyamoorthy, Adarsh Jagan and Weerakoon, Kasun and Manocha, Dinesh",
  title   = "{GA-Nav: Efficient Terrain Segmentation for Robot Navigation in Unstructured Outdoor Environments}",
  journal = "IEEE Robotics and Automation Letters",
  year    = 2022,
}

@InProceedings{ICRA:Gaussian,
  author    = "Abe Leininger and Mahmoud Ali and Hassan Jardali and Lantao Liu",
  title     = "{Gaussian Process-based Traversability Analysis for Terrain Mapless Navigation}",
  booktitle = "IEEE International Conference on Robotics and Automation (ICRA)",
  year      = 2024,
}

@InProceedings{RSS:FTE,
  author    = "Jonas Frey and Matias Mattamala and Nived Chebrolu and Cesar Cadena and Maurice Fallon and Marco Hutter",
  title     = "{Fast Traversability Estimation for Wild Visual Navigation}",
  booktitle = "Robotics: Science and Systems (RSS)",
  year      = 2023,
}

@InProceedings{ICRA:STEPP,
  author    = "Sebastian Ægidius and Dennis Hadjivelichkov and Jianhao Jiao and Jonathan Embley-Riches and Dimitrios Kanoulas",
  title     = "{Watch Your STEPP: Semantic Traversability Estimation using Pose Projected Features}",
  booktitle = "IEEE International Conference on Robotics and Automation (ICRA)",
  year      = 2025,
}

@InProceedings{ICRA:GSAT,
  author    = "Dongjin Cho and Miryeong Park and Juhui Lee and Geonmo Yang and Younggun Cho",
  title     = "{GSAT: Geometric Traversability Estimation using Self-supervised Learning with Anomaly Detection for Diverse Terrains}",
  booktitle = "IEEE International Conference on Robotics and Automation (ICRA)",
  year      = 2026,
}

@InProceedings{ECMR:FtF,
  author    = "Khizar, Zaar and Laconte, Johann and Lenain, Roland and Aufrère, Romuald",
  title     = "{Feeling the Force: A Nuanced Physics-based Traversability Sensor for Navigation in Unstructured Vegetation}",
  booktitle = "European Conference on Mobile Robots (ECMR)",
  year      = 2025,
}

@InProceedings{ICRA:GND,
  author    = "Liang, Jing and Das, Dibyendu and Song, Daeun and Shuvo, Md Nahid Hasan and Durrani, Mohammad and Taranath, Karthik and Penskiy, Ivan and Manocha, Dinesh and Xiao, Xuesu",
  title     = "{GND: Global Navigation Dataset With Multi-Modal Perception and Multi-Category Traversability in Outdoor Campus Environments}",
  booktitle = "IEEE International Conference on Robotics and Automation (ICRA)",
  year      = 2025,
}

@InProceedings{ICRA:V-STRONG,
  author    = "Sanghun Jung and Joonho Lee and Xiangyun Meng and Byron Boots and Alexander Lambert",
  title     = "{V-STRONG: Visual Self-Supervised Traversability Learning for Off-road Navigation}",
  booktitle = "IEEE International Conference on Robotics and Automation (ICRA)",
  year      = 2024,
}

@InProceedings{EMNLP:SIA-VLN,
  author    = "Yicong Hong and Cristian Rodriguez-Opazo and Qi Wu and Stephen Gould",
  title     = "{Sub-Instruction Aware Vision-and-Language Navigation}",
  booktitle = "Conference on Empirical Methods in Natural Language Processing (EMNLP)",
  year      = 2020,
}

@article{RAL:DILLM-VLN,
  author  = "Jiawei Wang and Teng Wang and Wenzhe Cai and Lele Xu and Changyin Sun",
  title   = "{Boosting Efficient Reinforcement Learning for Vision-and-Language Navigation With Open-Sourced LLM}",
  journal = "Robotics and Automation Letters",
  year    = 2025,
}


\end{document}